\documentclass{article}

\usepackage{arxiv}

\usepackage[utf8]{inputenc} 
\usepackage[T1]{fontenc}    
\usepackage{hyperref}       
\usepackage{url}            
\usepackage{booktabs}       
\usepackage{amsfonts}       
\usepackage{nicefrac}       
\usepackage{microtype}      
\usepackage{lipsum}		

\usepackage[numbers]{natbib}
\usepackage{graphicx}
\usepackage{amsmath}
\usepackage{mathptmx}

\title{A Regression Framework for Predicting User's Next Location using Call Detail Records}

\author{
  Mohammad Saleh Mahdizadeh \\
  School of Electrical and Computer Engineering\\
  University of Tehran\\
  North Kargar, Tehran, Iran \\
  \texttt{mahdizadeh.s@ut.ac.ir} \\
   \And
 Behnam Bahrak \\
   School of Electrical and Computer Engineering\\
  University of Tehran\\
  North Kargar, Tehran, Iran \\
  \texttt{bahrak@ut.ac.ir} \\
}

\begin{document}
\maketitle

\begin{abstract}
With the growth of using cell phones and the increase in diversity of smart mobile devices, a massive volume of data is generated continuously in the process of using these devices. Among these data, Call Detail Records, CDR, is highly remarkable. Since CDR contains both temporal and spatial labels, mobility analysis of CDR is one of the favorite subjects of study among the researchers. The user next location prediction is one of the main problems in the field of human mobility analysis. In this paper, we propose a data processing framework to predict user next location. We propose domain-specific data processing strategies and design a deep neural network model which is based on recurrent neurons and perform regression tasks. Using this prediction framework, the error of the prediction decreases from 74\% to 55\%  in comparison to the worst and best performing traditional models. Methods, strategies, the framework and the results of this paper can be helpful in many applications such as urban planning and digital marketing.
\end{abstract}


\section{Introduction}
Location prediction is a problem in the area of human mobility analysis which has drawn the attention of data scientists and machine learning researchers to itself during the past decade. Despite the simplicity of the problem description, the solution is expected to be complicated due to its dependence to many factors such as the environment, variability in users’ habits and their locations, and different formats of the data.

Location prediction problem could be described using the available data for users’ locations and the environment in which we need to predict the next locations of these users. Such data usually contain historical records that include information about time, location, user characteristics, and the infrastructure that collects the location information.

Prediction of human mobility has a wide variety of applications and many real-life use cases. Knowledge about the places where each user may traverse through would supply valuable information to use in urban planning, country infrastructure management, and public transportation organization. Smart advertising and businesses would hugely benefit from such predictions. There are also many different domains like social and cultural researches, disease epidemiology studies, criminology investigations, cellular network infrastructure administration, etc. that could benefit from the information provided from human mobility predictions \cite{blondel2015survey}.

With the growth of smart mobile devices that are spread geographically apart in the environment, gathering human mobility data has become much easier compared to the traditional methods like surveys or census which suffer from the static or low resolution spatial and temporal information. There are various sources of data used by researches in the field of location prediction. For example, Lenormand et al. \cite{lenormand2014cross} has done a cross-check analysis by comparing results obtained from different sources of data (cell phones, twitter and census) and compared the levels of correlations between these sources in three different aspects of spatial distribution of population, temporal evolution of people density, and mobility patterns of individuals.

The most common data resource that has been used in human mobility researches is mobile devices. These data might be collected from GPS-based navigation applications, social networks which work over the internet, or cellular networks infrastructure.

Cellular networks possess valuable data like call detail records (CDRs) of the users, accounting information, and infrastructure information which can play an effective role in human mobility analysis. CDR is a type of metadata which describes users’ activities in a cellular network. CDR data are commonly used for the purpose of billing users, value-added services, and network maintenance and optimizations by cellular network operators and infrastructure maintainers. However, having both spatial and temporal information about users has also made CDR a good resource for analyzing human mobility.

In this paper, we propose a framework which consists of a recurrent neural network regression model for predicting users’ next location based on the spatio-temporal information in CDR records. What distinguishes this solution from similar models proposed for location prediction is:  (1) it uses domain-specific data preparation methods to form meaningful user trajectories to make them easier to process by the proposed model, and (2) it uses geographic coordinates instead of domain-independent semantic labels for locations in the process of learning and prediction.

In order to compare the performance of our proposed framework with existing models, we implemented two baseline models and a common recurrent neural network classification model that have been widely used in the field of location prediction. We test these four models/frameworks on a real-world dataset, which includes CDR data collected from 12 users over a period of two years by one of the largest mobile phone operators in Iran, and compare the results with respect to different metrics of performance.

The remainder of this paper is organized as follows. We first provide an overview of the related works in the field of human mobility prediction in Section \ref{sec:relatedworks}. Section \ref{sec:datasetdescription} introduces the dataset and Section \ref{sec:problemformulation} formulates the problem. In section \ref{sec:traditionalmodels}, we discuss traditional models and  in Section \ref{fig:framework} we introduce our prediction framework. We explain the experiment results in section \ref{sec:experiment}, and finally Section \ref{sec:conclusion} concludes the paper.

\section{Related Works}
\label{sec:relatedworks}
Song et al. have explored limits of predictability in human mobility and by measuring the entropy of users’ visited locations found a 93\% potential predictability across their dataset which includes 50,000 individuals’ mobility data in 3 months collected by mobile carriers for billing purposes \cite{song2010limits}. It is also showed that this limit of predictability is not much variable for users with different distance coverage. 

Leng et al. \cite{leng2016urban} proposed a recurrent neural network model for next location prediction on a large-scale CDR dataset collected from tourists in Andorra. The model was inspired by Natural Language Processing (NLP) solution models to predict the next cell tower of the users’ presence. It uses a strategy of mapping each cell tower to a word and then converting the sequence of visited locations to sentences. The proposed model is a four-layered recurrent neural network consisting of an input layer, an embedding layer, an LSTM layer, and a dense output layer which receives a sequence of locations as input, and outputs the prediction of the next location that user might visit. On a limited state space of 100 static cell towers, their model achieves a prediction accuracy of 67\%.  They have also used DBSCAN clustering method to cluster the state space of the cell towers and decrease the possible locations to 25 regions (instead of 100), which increases the accuracy of the predictions.

Gomes et al. \cite{gomes2013will} proposed a general framework which uses contextual data in addition to spatial and temporal data to increase the accuracy of the prediction. The framework is also capable of online learning and prediction, and its core can be integrated with an anytime classification model that learns incrementally, but it does not propose any specific prediction algorithm. Results of their experiment on Nokia MDC dataset show that the prediction capability of the model varies among the users with respect to their mobility behavior.

In \cite{karatzoglou2018seq2seq}, a Seq2Seq learning approach is proposed that uses attentional recurrent neural networks. It is showed that using attention-based Seq2Seq learning on users’ semantic trajectories could improve the accuracy of users’ location prediction. The proposed model, DeepMove, outperforms a simple Markov model and a general recurrent neural network (RNN) model by more than 10\%.

Cuttone et al. \cite{cuttone2018understanding} discussed the factors that may affect the performance of human mobility prediction and cause a wide variety of accuracies for the prediction models proposed in the literature. These factors include temporal and spatial resolution of data, new place transitions, frequency of visiting different locations, etc.

Although there are various researches which proposed methods for predicting users’ next locations based on patterns or models learned from large and accurate temporal and spatial datasets, some drawbacks are notable in many of these researches. First, many of the proposed methods, models or frameworks are fed only with sequential semantic data consisting of discrete location IDs instead of continuous coordinates. Geographic coordinates are more informative and provide important details about the distance between locations, patterns of trajectories on the coordinate plane, and geographic relevance of two adjacent locations. By removing geographic coordinates from location trajectories of the users and using domain-independent semantic labels, we lose an important source of information.

Another common drawback of these researches is their focus on the model, and not paying attention to the data and its properties. Preparing data appropriately in order to feed into the model is as important as the model design, especially on sparse and heterogeneous datasets. To compensate for the sparsity and heterogeneity in the data, we need to have a careful data preparation phase. CDR records are usually produced when a user makes or receives a call, and the irregularity and sparsity of calls reduces our collected information about users’ trajectories. If the data prepared improperly and the structure of the model’s input is poorly designed, ignoring the quality and quantity of the data, even the best existing models will provide weak and imperfect results.

\section{Dataset Description}
\label{sec:datasetdescription}
Call Detail Records (CDR) is a collection of records consisting of information about users’ activity in cellular networks. CDRs are usually used for the purpose of billing, cellular infrastructure maintenance, and resource management by service providers.

A CDR often includes source and destination phone number, date and time, base transceiver station (BTS) IDs, device information of the parties, type of the communication (call, message, data packets, etc.), duration, and network operator IDs of every communication over a cellular network. Every CDR record is initiated when a communication occurs and filled out as soon as the communication ends. Thus each CDR record contains metadata about a successful communication over the cellular network. Because of the usual sparsity and irregularity in communications of most users and heterogeneity of network devices, CDR is considered a sparse and erratic source of information which requires careful cleaning and preparation compared to high resolution, regular, and dense datasets such as network devices’ logs.

In CDR, the exact time and the date of the communication are available and we can use them as temporal tags. Spatial data in a CDR may not be directly available in the format of the geographical coordinates (latitude and longitude).  Usually the location data in a user’s CDR is the identification code of its serving BTS which includes Location Area Code (LAC) and Cell ID (CELL). The LAC and CELL information along with the mobile network code (MNC) can be uniquely mapped to the BTS’s geographic coordinates.  Since the coverage area of modern BTSes in densely populated urban areas is very limited, we can approximate the location of the user by the location of its serving BTS.

In CDR, spatial data forms trajectories for each user in the form of location sequences. These trajectories could be in the format of semantic location label sequences or geographic location sequences. Semantic trajectories are commonly in the type of cell tower unique IDs (LAC and CELL). Geographic location sequences are arrays of 2D vectors which contain latitude and longitude of those locations.

In semantic location sequences, the trajectories could be treated as arrays of independent labels which could be interpreted as location classes. The state space of these classes is the collection of all cell towers in the dataset, and the whole sequences should be processed by classification models/frameworks. On the other hand, geographic location sequences form an array of real valued vectors which is not treatable as independent classes, but should be interpreted and processed by regression models/frameworks which takes real-valued inputs and gives real-valued outputs.  We will talk about the drawbacks and benefits of each type of these sequences in Section \ref{sec:clsprep} and Section \ref{sec:regnet}.

The dataset used in this research is provided by one of the largest mobile phone operators in Iran. The dataset includes thousands of anonymized CDR records for 12 users who live in urban areas through 1.5-3 years timespans. The records include both voice calls and short messages of these users. The dataset variables used in this paper are UserID (anonymized phone number of users), serving BTS identifications (LAC and CELL), country and operator identifications (MCC and MNC), and the timestamps of the records (Date and Time).

\section{Problem Formulation}
\label{sec:problemformulation}
Location prediction problem is about getting information of the users and the history of their trajectories and propose a location or a sequence of locations as the prediction of next places that users may visit in the time ahead. This information might contain temporal and spatial labels of the visited cells, points of interest, frequencies of the visited locations, the gap time between adjacent recorded locations, etc. 

Each user’s historical location sequence can be formulated as a sequence like $(L_1, L_2, L_3, \dotsc, L_n)$ where $L_i$ is a structure that contains temporal and spatial labels of location $i$ which user visited before the location $i-1$ and after the location $i+1$. For example, $L_j$ could be expressed as $\{t, l, lat, lon\}$. $t$ attribute is the temporal label of the location record. $l$ is a unique independent ID for each visited BTS which is created from concatenating CELL and LAC attributes. $lat$ and $lon$ also specify the geographic coordinate (i.e. latitude and longitude) of the visited location. For example, If we want to show the latitude of the location $i$, we will use the notation $lat_i$.

In this paper, we propose a framework to predict next location of a user. The framework takes the sequence $(L_1, L_2, \dotsc, L_n)$ as input for a user and predicts $[lat_{n+1}, lon_{n+1}]$ which is a vector that contains geographical coordinates of the predicted location. The model used in our proposed framework is a regression model.

\section{Traditional Models}
\label{sec:traditionalmodels}
In order to compare the results of our proposed framework with the existing and traditional models that have been used to predict the next location of a user, in this section we introduce these models thoroughly.

\subsection{Most Frequent Next Visited Location Model}
The first model that we discuss is a naïve model that predicts the next visiting location of a user based on the most frequent location that he/she has visited after the current location. This model uses the history of trajectories of the user and creates a frequency table for every known location (cell) in the history of the user. In each frequency table (that belongs to a specific location), there exist all next locations that has been visited after the specified location along with their frequencies.

When the model wants to make a prediction for the next location of the user in a specific location, it refers to the frequency table of that location for the user and picks the most frequent location that has been visited in the history of the user’s trajectories and offers it as the predicted next location.

\begin{figure}[h!]
	\centering
	\includegraphics[width=0.95\textwidth]{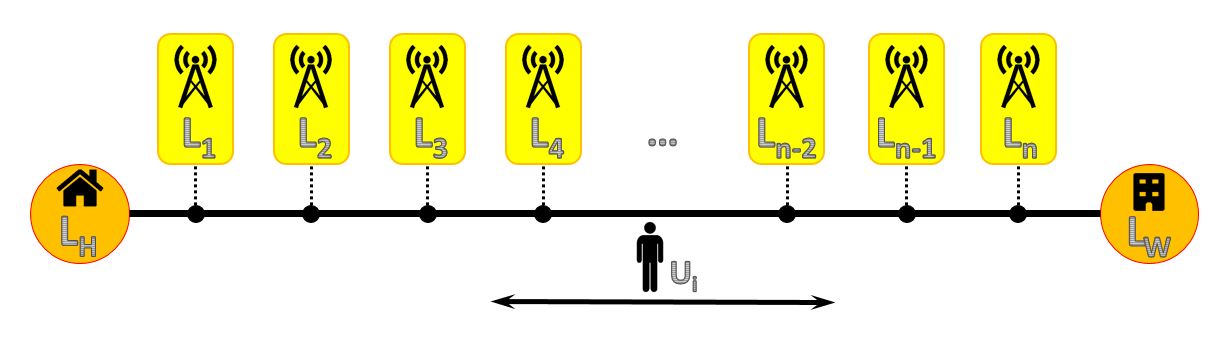}
	\caption{An example of a user's daily mobility.}
	\label{fig:userpattern}
\end{figure}

This model, despite the simplicity of the design and implementation, has a reasonable estimation of the mobility behavior of users. For example, assume that a user (e.g. $U_i$) has a typical mobility pattern moving from his/her home location $L_H$ to his/her work location $L_W$. Through traversing this typical route, user $U_i$ may visit some intermediate cell towers as well, e.g., $(L_1,L_2,L_3,L_4, \dotsc, L_n)$ in the order of visiting from home to the work. Figure \ref{fig:userpattern} shows the daily mobility scheme of $U_i$.

\begin{figure}[h!]
	\centering
	\includegraphics[width=0.95\textwidth]{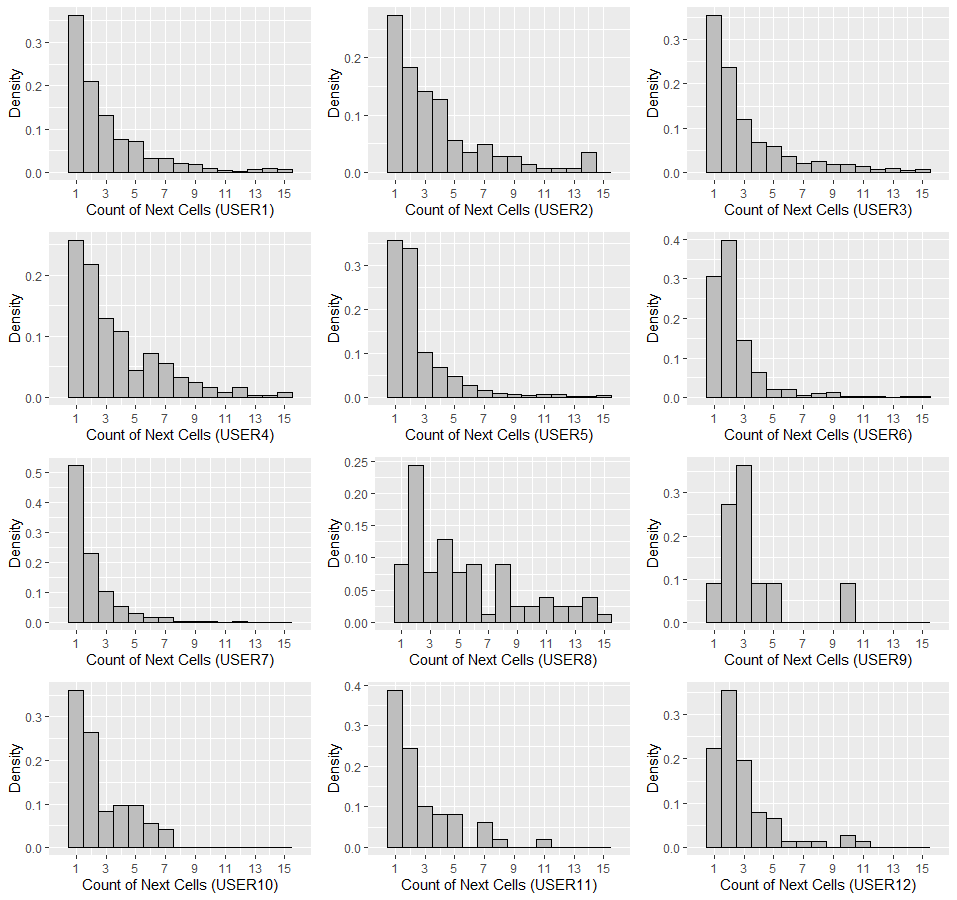}
	\caption{Distribution of number of next locations for users in the dataset.}
	\label{fig:nextcellsdist}
\end{figure}

The high accuracy achieved by this model could be justified in two cases. First, if any location has only one subsequent location in the historical trajectories of the user, that location would be selected as the most frequent location, and the prediction for that next location by the naïve model would be correct. There exist lots of locations in the dataset that are always followed by one specific visited location, and that portion of the data would be assessed as correct prediction by the naïve model. Figure \ref{fig:nextcellsdist} shows the distribution of number of next locations for 12 users in our dataset.

Another reason that could explain the acceptable performance of this model is the repetitive nature of human mobility itself. Most people follow deterministic traversing routines during the weekdays.

Considering these reasons, it is not a surprise that this naïve model gains high accuracies in many cases. This model extracts the main pattern of mobility with a simple process, and for the users with repetitive trajectories, this works well.

\subsection{Markov Chain Decision Model}
Markov chain decision model is another popular model used by a wide variety of human mobility researches to solve the problem of individual location prediction \cite{gambs2012next, mathew2012predicting, asahara2011pedestrian}. Markov models build the transition graph of the locations based on the user’s traversed trajectories. After extracting transition probabilities based on location sequence of a user, this model proposes a location that has the highest probability among the probable next locations in respect to the input location sequence.
The transition matrix could be created from the historic location sequence of the user. The model first makes independent states, where each state maps to an individual location label, then calculates the probabilities based on the historical frequencies of the possible visited locations after each state.

The most frequent next visited location model is a simple form of Markov Chain Decision Model, considers only the last visited location instead of a sequence of locations. This does not mean that Markov Chain Model always perform better than the most frequent next visited location model.  Our experiments (Section \ref{sec:experiment}) show that the naïve model outperforms Markov Chain model in many cases.

\subsection{Recurrent Neural Network Classification Model}
Researches in the field of artificial neural networks have been drastically progressed in recent years. Passing through classic feed-forward neural networks, powerful deep and recurrent neural networks have enabled reaching targets that had been seemed to be unreachable before.

Simple recurrent neural networks are similar to classic feed-forward neural networks, with the difference of existing a connection from recurrent layers to themselves. These loopback connections make possible of modeling data that is sequenced or time-dependent. This feature of recurrent neural networks has made them one of the best solution for modeling and predicting sequences.

Since classic RNN nodes suffer from drawbacks such as vanishing gradients, Long Short-Term Memories (LSTM) are developed to address these problems and replace classic RNNs. LSTM design includes memory cell units that can maintain long term memories of the input sequences. There exists a set of gates that control the input information and choose the flow and the volume of it through internal units. This architecture makes it possible to memorize long and short pattern over the sequence of data.

Since finding the future location of a user is a type of sequence prediction problem, and considering the power of recurrent neural networks in predicting sequences, the question of using WHAT method is changing to HOW to use the method. To solve the location prediction problem efficiently with RNNs, we have to address problems such as data formatting, network design, and parameter tuning.

Here we briefly review the common architecture that has been used widely to predict sequences. In addition to predicting human mobility, this model has a variety of applications in natural language processing (NLP), \cite{leng2016urban, alahi2016social, Langford2017}.

\begin{figure}[h!]
	\centering
	\includegraphics[width=0.6\textwidth]{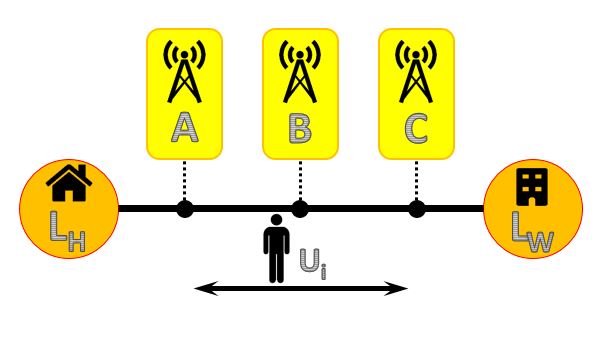}
	\caption{An example of a user's daily mobility with the trajectory size of 5.}
	\label{fig:userpatternshort}
\end{figure}

The Next Location Prediction problem could be stated as a classification problem. While each location could be expressed as an independent class, the model should classify a sequence of input classes into the result class. For example, assume $U_i$ has a trajectory length of 5 in the past day, going from home to work and vice versa, like the path showed in Figure \ref{fig:userpatternshort}. The trajectory could be stated as the sequence $(L_H, A, B, C)$ and the expected outcome is $L_H$. So, the designed model would take a vector of classes in the input and classify the input into one of the classes $\{A, B, C, L_H, L_W\}$. In this formulation of the problem, the classes (location labels) are independent and there exists no relationship between the classes.

\begin{figure}[h!]
	\centering
	\includegraphics[width=0.95\textwidth]{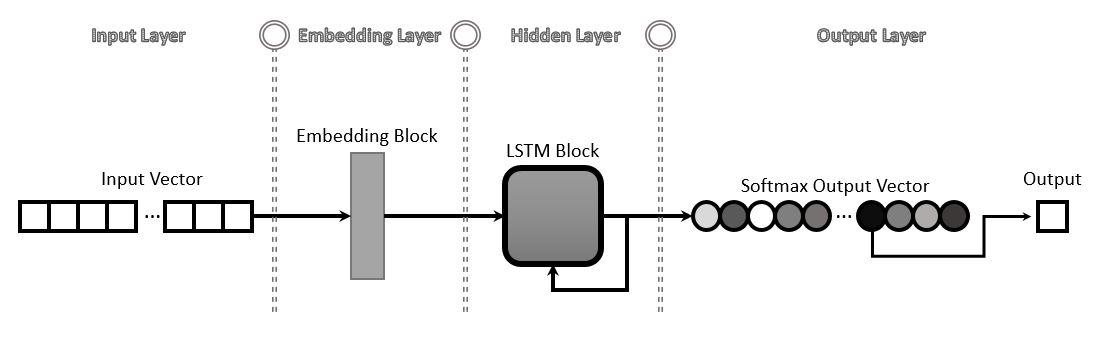}
	\caption{Recurrent neural network classifier architecture.}
	\label{fig:clsnet}
\end{figure}

The architecture of most common LSTM recurrent neural network classification model for this type of formulation is showed in Figure \ref{fig:clsnet}. This model is a four-layer recurrent network which takes one-hot-encoded vectors in the input at each round. This input feeds the next layer, which is an embedding layer. The embedding layer takes vectors of one-hot-encoded class and transform them into the real-valued vectors.

The third layer consists of LSTM nodes and accepts real-valued vectors of the embedding layer. In contrast to the classic RNN nodes, these nodes do not suffer from the vanishing gradient effect or the exploding gradient effect \cite{bengio1994learning}, and also they can keep up information from a far distance in the time, keeping up the effect and dependency of far inputs \cite{hochreiter1997long}. These neurons try to fit a complicated model on the input vectors with respect to their distribution in time.

The last layer is a categorical classification layer which uses the output of the LSTM layer. The activation function of this layer’s nodes is softmax (Equation \ref{eq:softmax}) and the loss function is cross entropy (Equation \ref{eq:crossentropy}). It is typical to use these functions when dealing with a multi-category neural network classifier. Softmax function calculates the probabilities of each target class, and the class with the highest probability would be considered as the predicted class. Cross entropy also computes the distance between the output distribution of the model and the true distribution. By applying the argmax function over the output layer, the result class would be determined.

\begin{equation}
\label{eq:softmax}
f(s)_i = \dfrac{e^{s_i}}{\sum_{j=1}^{C} e^{s_j}} \quad \textrm{for} \quad i \in \{1, \dotsc, C\}
\end{equation}

\begin{equation}
\label{eq:crossentropy}
CE = - \sum_{i=1}^{C}t_i\log(f(s)_i)
\end{equation}

Note that since each location has a distinct class, the number of the output layer neurons is equal to the number of known locations in the history of the user’s trajectories.

\subsubsection{Data Preparation}
\label{sec:clsprep}
As mentioned above, the recurrent neural network classification model takes a sequence of user’s traversed locations and predicts a location as the result. The input sequence and the output location are both in one-hot-encoded format.

The length of the input sequence is an important factor because the whole history of the user’s trajectories could not be fed into the model, so it has to be divided into the smaller chunks. Long input sequences result in small training sets which hinder the learning process. Also with long sequences, the classification process becomes more complicated.

\begin{figure}[h!]
	\centering
	\includegraphics[width=0.95\textwidth]{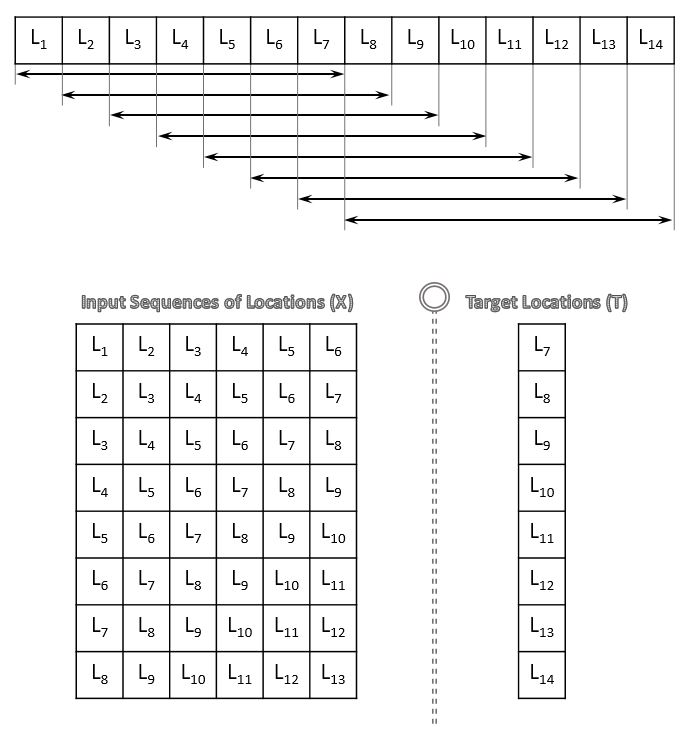}
	\caption{An example of dividing user's traversed sequence into smaller subsequences.}
	\label{fig:prep1}
\end{figure}

There are many strategies to slice the user’s trajectories into the smaller sequences. The most typical method that has been widely used is applying a constant-length window and shifting it over the sequence step by step. Assume a fixed window with the length of $w$. If we shift this window from the first element of the user’s traversed sequence of length $n$, we obtain $n-w+1$ subsequences. Now we can consider each of these subsequences as a trajectory that could be fed into the recurrent neural network. Each input sequence starts from $i$ to $i+w-2$ and the target location would be $i+w-1$ while the value of $i$ is shifting from $1$ to $n-w+1$. Figure \ref{fig:prep1} shows an example of dividing a user’s traversed sequence into smaller subsequences.

Another data preparation technique that we use is removing unknown locations (i.e. locations that the user has not visited before and are not included in the train data) from the test set of the data. The design of the model depends on the number of the known locations in the locations state space, e.g. the number of nodes in the last layer of the neural network is equal to the number of possible locations. So, in the process of prediction over the test data, the model cannot deal with new locations that were not seen in the train data. Various studies neglected the disadvantage of classification models in the prediction of human mobility. Many of these researches have done their experiments with the assumption of knowing the whole location space, which is not a practical assumption.

It is also possible to cluster locations with respect to their distances, and instead of predicting individual location, predict clusters of locations. Spatial clustering procedure needs the geographic coordinates information (latitude and longitude) of the locations, possibly supplied from CDR dataset or independent location API services such as OpenCellID\footnote{OpenCellID by Unwired Labs: \it{http://www.opencellid.org}}. This clustering could be done using clustering algorithms like DBSCAN \cite{leng2016urban}. Spatial clustering would increase the accuracy of the model because decreasing the size of the location space, increases the probability of choosing the correct location.

Like every other classifier, the performance of our LSTM classification model could be stated by the value of the accuracy, i.e. the ratio of correctly classified instances. The performance could also be measured by the average prediction error which is the average geographic distance between the real and the predicted locations of a user. This value is also similar to the amount of loss that the neural network trains to the matter of that. Also, it could be stated by the value of mean square error (MSE) or mean absolute error (MAE), but the most real-world sensible value would be the average geographical distance between outcome and expectation, mainly calculated with the help of haversine formula in the domain of geographical axis.

\section{The Framework}
\label{sec:framework}
Our framework consists of 5 units. The whole framework receives new updates in the format of CDR data. At the first step, raw CDR data are processed using the Cleaning Unit. This unit performs primary and generic cleaning of the data in order to fix common flaws that may exist in the available raw CDR data. The cleaning unit then pass the cleaned data to the Profiling Unit. This unit separates the cleaned CDR records for each of the users, add the extension labels like $l$, $lat$ and $lon$ and then saves them into the user’s specific data pool. After this step, the process of the training begins for each of the users that receive new CDR updates. 

At the first step of the training for each of the users, the Preparation Unit fetches the data from the user’s pool. After that, the prepared data, which is a collection consisting of different trajectories, is normalizes and fed into the model. The model, which is an LSTM recurrent neural network regression model, is trained using the prepared data. After completion of the process of training, the trained model is stored for the user for later usage of the prediction. This whole process runs each time the new CDR updates are received.

\begin{figure}[h!]
	\centering
	\includegraphics[width=0.95\textwidth]{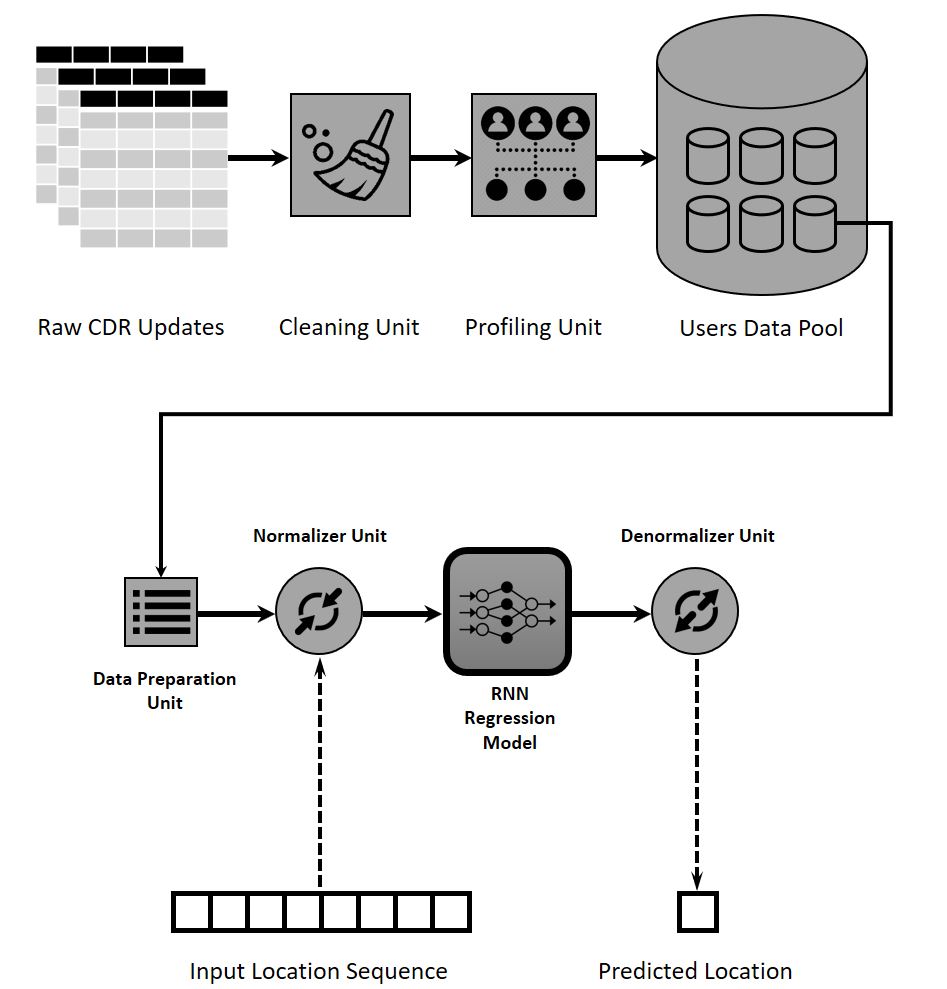}
	\caption{An overview of the proposed regression framework.}
	\label{fig:framework}
\end{figure}

When the framework is asked for a next location prediction for a user over the specific trajectory, the stored model for the user is called and the process of prediction starts. The result then denormalizes and places at the output. Figure \ref{fig:framework} shows the schematic of the whole proposed framework. In the following, we will discuss each of the framework’s units specifically.

\subsection{Data Cleaning and Profiling Units}
CDR data which are collected from cellular network service providers would need basic cleanings in order to remove erratic data and malformations. Raw CDR data typically has the following issues that must be fixed and cleaned, before we use it in our predictive model:
\begin{itemize}
	\item 
	Fully duplicate records that usually caused by the huge flow of call events gathered from different devices all over the network in a short period of time. It is a good practice to remove fully duplicate records at the cleaning phase, because this leads to lower overhead and decreases the volume of the process for the next phases. Also removing duplicate records, prevents computing false frequency for visited locations of a user.
	\item 
	Not Available values (NAs) that ruin the spatio-temporal dimension of the location sequences. If this unavailability damages the structure information of an $L_i$, the $L_i$ should be omitted from the location sequence. 
	\item 
	Dissimilarities in a user’s identification. Sometimes a phone number (which we use as user ID) may appear in different shapes. For example, in some CDR records the phone number may include the country code, but in other records it simply starts with the area code. Before doing any analysis on the dataset, we have to unify IDs of different format which are the same.
\end{itemize}

\subsection{Recurrent Neural Network Regression Model}
\label{sec:regnet}
It is a common practice for researches in the field of human mobility prediction to use classification models. This is mostly because often geographical coordinates of users’ trajectories are not available and we have to work with location labels. Another reason would be the complicated process of data formatting and preparation to form the regression models in compare to the classification models. 

Regression models usually require a more complicated design process. The problem of location prediction could also be solved by regression models because the data could be represented by real-valued geographical coordinates. In this section we propose a regression model to predict the coordinates of the next location of a user. 

One advantage of our proposed model is preserving the distance between locations. In contrast to classification models where the distance between model’s output and target classes is not considered, if the regression model’s output is different from the target value, the geographical distance between the outcome and the target would be interpreted as error. This results in a more practical approach for predicting next location of a user. For example, assume a user is located in the coverage area of two adjacent BTS. If the user equipment connects to BTS1 and our classification model predicts BTS2, since the label of these BTSes are different, we count this as a misclassification. But in practice predicting BTS2, because of its closeness to BTS1 is a good enough output. To address this problem some researches perform clustering before classification to unify the BTSes that are close together, but this decreases the resolution of the data.

Another advantage of the regression model in comparison to the classification model would be the ability to interpret new locations that have not appeared in the trajectories of the users in the past. In real-world experiments, we observed that the problem of dealing with new locations and the effect of variability in the locations state space could be significant.

Here, we propose a recurrent neural network regression model to predict the next location of a user. This model is composed of 4 layers. The first layer is the input layer, accepting real-valued vectors of pairs representing points on the two-dimensional surface. These values are derived from the normalization process performed by the normalizer unit of the framework that will be explained later.

The second and third layers are composed of LSTM layers; each one creates a level of complexity in the process of model formation. The result of the LSTM layers transfers to the last layer, which is a dense layer consisting of fully connected neurons which create output by filtering the result through an activation function. The output is a vector of length 2, which represents the predicted coordinates produced by the network. This outcome should be then denormalized using the denormalizer unit in the framework to create a 2D vector that includes latitude and longitude of the predicted location.

The performance of this regression neural network model highly depends on the tuning of the hyper-parameters and formatting of the input data. In the following section, we discuss the data preparation methods that help us to gain high performances using this model.

\subsubsection{Data Preparation Unit}
As mentioned in previous section, the whole historical sequence of a user’s trajectories could not be fed into the network and it should be divided into smaller subsequences of trajectories. Here, we discuss four methods for deviding a user’s traversed trajectories into smaller pieces. 

\begin{figure}[h!]
	\centering
	\includegraphics[width=0.7\textwidth]{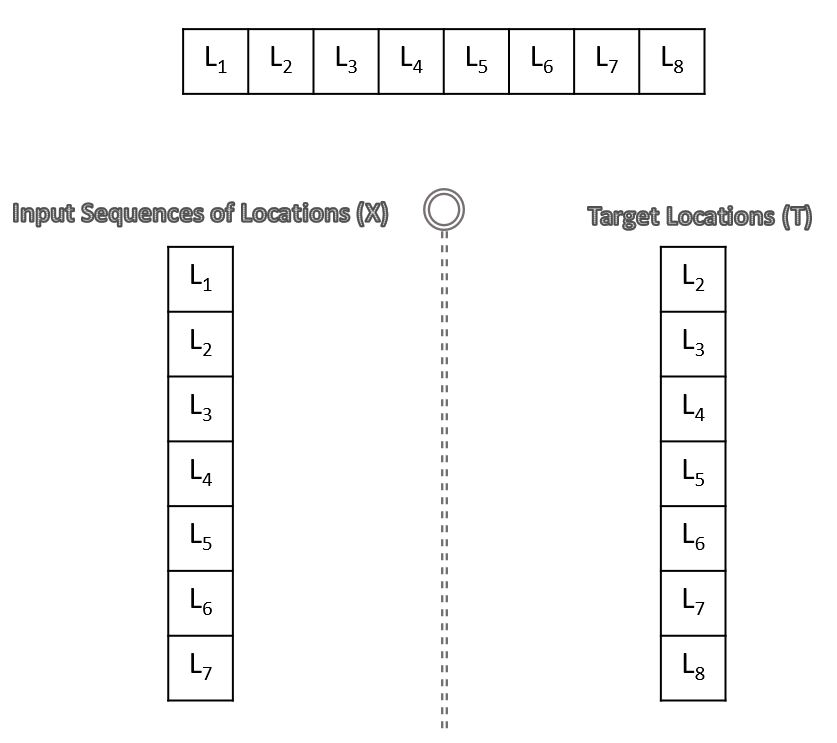}
	\caption{An example of data preparation using the first method.}
	\label{fig:prep0}
\end{figure}

In the first method, every input sequence would have a length of 1. In this formatting, the sequence of the user’s trajectories is transformed into a two-column table, the first column is the input of the network ($X$) and the second one is the expected result that the neural network learns ($T$). Figure \ref{fig:prep0} shows an example of data formatting using this method.In this method, the prediction only performed using only the last location of the user at each time. Using this method, we should  use non-random batching in the process of training the network, or the relationship between adjacent records would be lost and the performance decreases drastically.

The second method considers a fixed-width window with the length of $w$ shifting over the whole sequence of the locations step by step. In this method, each record of $X$  consists of the last $w$ locations from each time step, and the present location would be the value of the expected result.  After processing a sequence with the length of $n$, this method creates $n-w+1$ records. An example of using this method was shown in Figure \ref{fig:prep1} in Section \ref{sec:clsprep}. Longer window results in longer training time for the neural network model, but this does not necessarily improve the performance. Long training sequences result in an overfitted model, and short windows diminishes the relationship among elements of the sequences and result in low accuracies. 

In the third method, we consider the mobility behavior of users instead of processing the data blindly. This method uses dynamic sizes for sequences of each record. Feed-forward neural networks cannot process dynamically sized inputs, but in recurrent neural networks are capable of processing inputs with dynamic sizes which allows us to generate meaningful sequences for each record. Like the first two methods, this method divides the user’s location sequence into smaller subsequences, but these subsequences are more meaningful. The measure of division in this method is the distance between timestamps of consecutive locations in the sequence.

The sequence of the user’s visited places consists of location labels traversed by the user. Each of these visited places has a timestamp showing the exact time of the user’s presence on that location. The occurence of events in the time axis is irregular because the CDR data is only recorded when a network activity such as a voice call, or SMS for the user happens. If the user stays in a fixed location, the same place will frequently be registered in the location sequence but with different timestamps. These repetitive locations are redundant data and only inform us that the user does not move and stays in a single location. These redundant data makes a significant bias in the process of learning the model, thus providing the reason to remove the consecutive similar locations in the location sequence. To preserve the information about the duration of the user’s presence in a location, we remove all consecutive similar locations in the sequence except the first and the last one. This also let us divide the location sequence into meaningful subsequence of trajectories.

\begin{figure}[h!]
	\centering
	\includegraphics[width=0.7\textwidth]{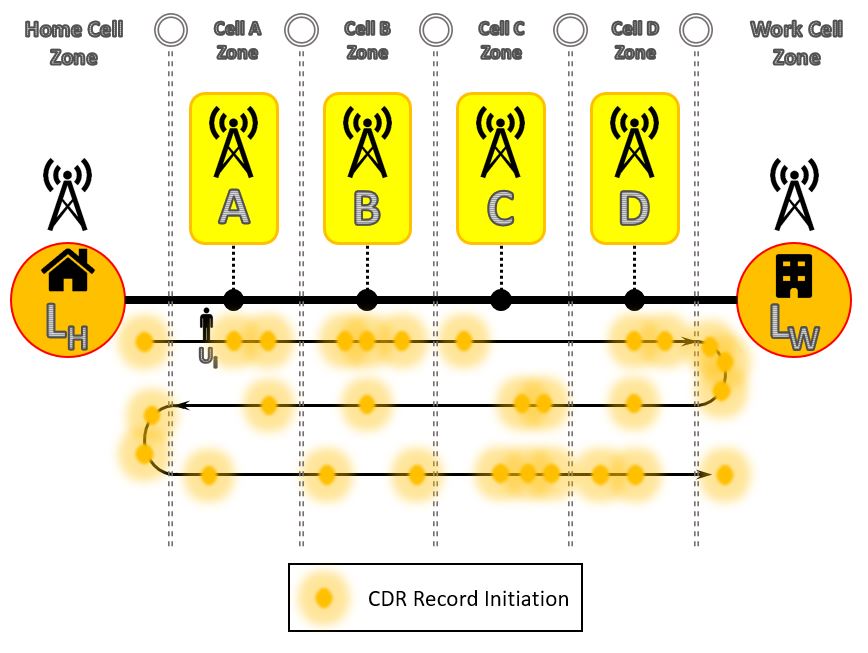}
	\caption{An example of a user's daily mobility.}
	\label{fig:userpatternlong}
\end{figure}

For example, assume user $U_i$ has an extreme behavior of only traversing through the path between home ($L_H$) and work ($L_W$) locations, also visiting the cell towers $\{A,B,C,D\}$ between these two locations. Figure \ref{fig:userpatternlong} shows the schematic of $U_i$’s traveling pattern.

\begin{figure}[h!]
	\centering
	\includegraphics[width=0.95\textwidth]{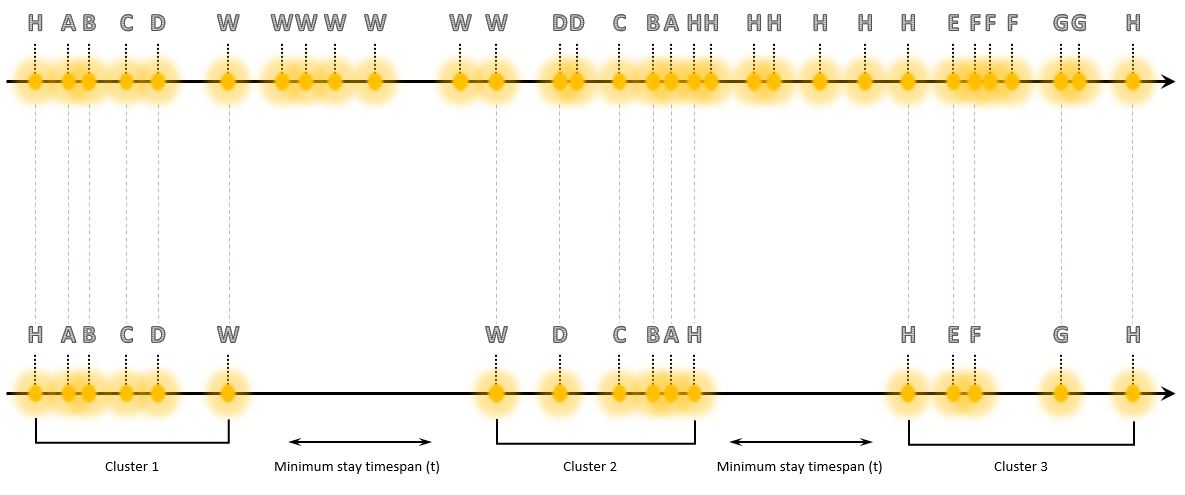}
	\caption{Extracting meaningful trajectories from the user's locations sequence.}
	\label{fig:prep2_1}
\end{figure}

Consider the locations sequence of the user in Figure \ref{fig:userpatternlong}. As you can see, this sequence has redundant data which is not related to the user’s mobility behavior. These redundant data cause many different sequences on two simple trajectory patterns of the $U_i$: $(L_H, A, B, C, D, L_W)$ and $(L_W, D, C, B, A, L_H)$. Since there exist many variants of these two simple mobility patterns, if we fed all these variants to the model, the model will not be able to summarize these patterns and extract the main patterns and learn them. As a result, the model performs poorly and its ability to learn decreases drastically which in turn increases the minimum error offset. Using the proposed method of data preparation, we remove redundant locations from the sequence, keeping only the first and the last occurrence of the adjacent identical locations with the constraint that the temporal distance between these two occurrences is more than a specific timespan like $t$. If the time difference between the first and the last occurrence of a location is less than $t$, we remove the last occurrence too. This specific timespan $t$, represents the minimum amount of waiting time to consider a user is settled in a place and not moving through. Thus, after applying the proposed method on the Ui traversed sequences, the adjusted sequence would become more meaningful. The next step is to divide the adjusted sequence into meaningful trajectories. It is possible to use a clustering algorithm like DBSCAN over the temporal axis to cluster the events, or even use a simpler strategy like slicing the sequence when the temporal distance between adjacent events is more than a specific value ($t$). Figure \ref{fig:prep2_1} shows the distribution of locations visits over the temporal axis and how clustering divides the whole sequence into the smaller meaningful subsequences. As a result, the model learns these trajectories and performs the prediction for similar trajectories much better.

The value of $t$, which represents the distance of slicing, should be adjusted independently for each user. In the experiment (Section \ref{sec:experiment}), we perform Grid Search over the value of $t$ and show that how the performance of the model changes with different values of this hyperparameter.

\begin{figure}[h!]
	\centering
	\includegraphics[width=0.6\textwidth]{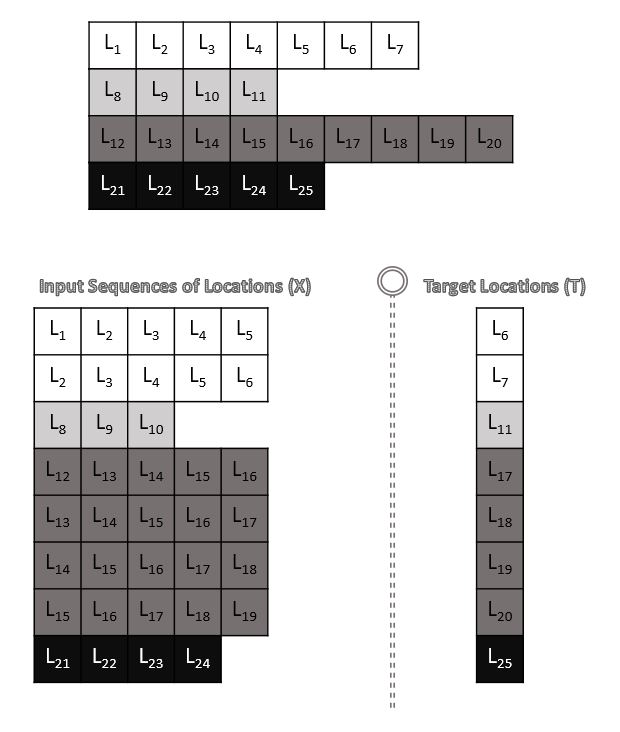}
	\caption{An example of data preparation using the hybrid method.}
	\label{fig:prep3}
\end{figure}

The fourth method is a hybrid of the second and the third methods. Using the third method on the user’s whole trajectories results in different subsequences with different lengths. There could be long subsequences generated by applying the third method, but the number of these long subsequences would not be large. These long subsequences could cause the model not to learn the input sequence properly, which in turn decreases the effect of these subsequences on the formation of the model. To solve this problem, we can apply the second method, i.e. constant-sized window, on the long sequences. This approach will break the long subsequences to subsequences with the maximum length of $w$, causing the model to interpret the long subsequences properly and increase their effect on the formed model. This method only applies fixed-width window on the subsequences longer than w and other subsequences stay intact. This increases the number of records which should be fed into the model and guarantees the maximum length of $w$ for the subsequences. An example of using this preparation method is showed in Figure \ref{fig:prep3}.

\subsubsection{Normalizer Unit}
Another necessary step in formatting the data for the recurrent neural network model is data normalization. The geographical coordinates are real-values, latitude ranges from $-90$ to $90$ and longitude ranges from $-180$ to $180$. The process of prediction is usually performed on the data of a user in a limited geographical area, e.g. a city. Thus, for each user, the values of latitudes and longitudes often do not vary more than $0.5$ unit for a large city and its rural areas. These values should be normalized for a recurrent neural network to help the network generate equalized weights for the features and improve the convergence rate and achieve lower training times \cite{laurent2016batch}.

Two conventional methods of data normalization are min-max scaling and variance scaling. In the min-max scaling method, all the data are transformed to the range of $[0,1]$ for absolute values using the Equation \ref{eq:minmaxscale}. This method transforms the smallest value to 0 and the largest one to 1. If we perform this method on the geographical data, all the points will move into the square with the area of 1 that is adjacent to the center of the axis. This normalization method is a standard procedure in many experiments. The main property of this method is the generation of positive normalized values  which forces us to use an activation function like ReLU for the last layer of the proposed regression model to avoid negative results by the recurrent neural network.

\begin{equation}
\label{eq:minmaxscale}
x^{\prime} = \frac{x - \min(x)}{\max(x) - \min(x)}
\end{equation}

Another method of normalization is variance scaling. In this method, each data point maps to a new point using Equation \ref{eq:varscale}. The normalized result will have zero mean and a variance of 1. In this method there would be negative results in the normalized values. Thus, the last layer of the recurrent neural network should have the linear activation function to allow the prediction results to have negative values as well.

\begin{equation}
\label{eq:varscale}
x^{\prime} = \frac{x -\bar{x}}{\sigma^2}
\end{equation}

\subsubsection{Training the Model}
Mini-batch training is a typical strategy to train neural networks. The conventional process of training neural networks is based on mini-batch stochastic gradient optimizations \cite{masters1804revisiting}. In general, it is not a good practice to use large mini-batches because the test error would increase \cite{masters1804revisiting}. Although we lose the available computations parallelism with small mini-batches, it is practical with today’s powerful hardware and computation devices \cite{masters1804revisiting}.

The common practice on choosing batches is to randomly pick records from the main data pool, or equivalently, picking mini-batches sequentially from shuffled training data. Although the randomization provides a more general model over the training data, it would also vanish any relations between adjacent training records. This would be a good strategy for many problems, but in the next location prediction problem, this might play a harmful role. As we discussed earlier, the prepared data contains a collection of sequences. The order of these sequences in the input is not random. For example, for $U_i$, if the input collection is $\{(A, B, C, D), (E, F, G)\}$, that means the $(E, F, G)$ trajectory would be probable to happen after the trajectory $(A, B, C, D)$. Thus, the trajectories in the input collection are not independent of each other. The main reason for using two layers of LSTM nodes in the proposed model is this hierarchy of trajectories in the input data; the first layer models the relation between adjacent sequences, and the second layer models the relation between adjacent locations in each of the sequences. Since we have to preserve the relations between adjacent sequences and adjacent locations, we should not pick the mini-batches randomly to keep the data in order.

Another method that we used in the process of training the model is preserving states of the LSTM nodes at each epoch of the training. In a stateless model, the states array of LSTM nodes would reset at each step of sequence processing. In contrast, in a  strategy, the previous states for each sample across the batches would propagate \cite{lstmstateful}. This propagation of LSTM nodes’ states among the batches would result in longer preserved relations among the collection of sequences and result in an increase of performance in problems with long-related and hierarchical sequences.

There exist two loss functions to choose in order to train the network, MSE or MAE. Although these two measures represent the loss similarly, MSE value would increase more than MAE in cases of scattered results, highlighting the high error values. 

\begin{figure}[p]
	\centering
	\includegraphics[width=0.95\textwidth]{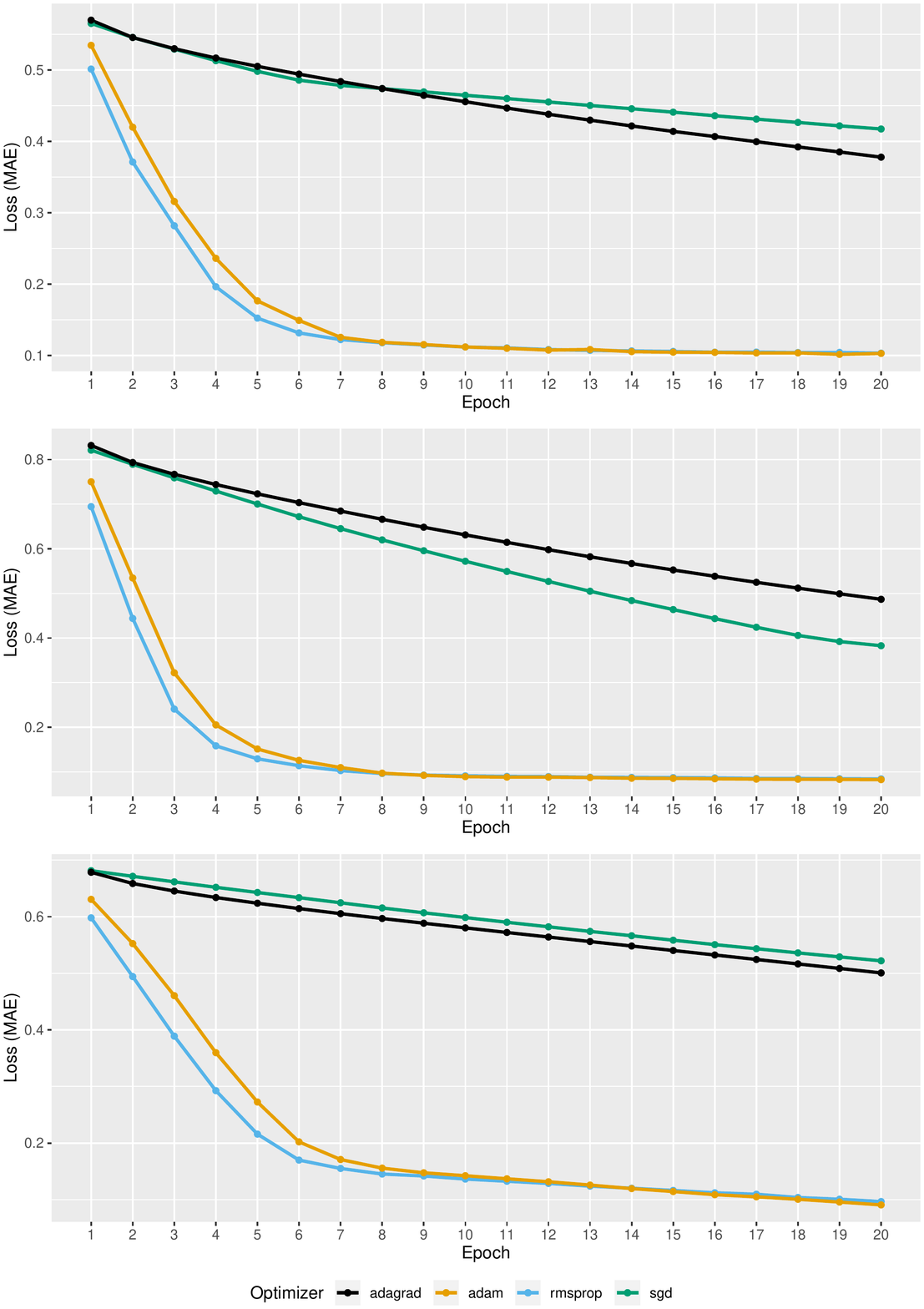}
	\caption{Variability of loss in the process of network training using four different optimizers for three sampled users.}
	\label{fig:optimizers}
\end{figure}

The process of training the neural network is performed using the Adam (Adaptive moment estimation) extension \cite{kingma2014adam}. Adam is an extension for Stochastic Gradient Descent (SGD) that combines the advantages of two other SGD extensions, AdaGrad \cite{duchi2011adaptive} and RMSProp \cite{Tieleman2012}. In addition to storing an exponentially decaying average of past squared gradients, Adam also keeps an exponentially decaying average of past gradients, similar to momentum \cite{ruder2016overview}. The authors in \cite{kingma2014adam} also show empirically that Adam works well in practice and compares favorably to other adaptive learning-method algorithms \cite{ruder2016overview}. In this experiment, we use default proposed learning parameters ($\beta_1 = 0.9$, $\beta_2 = 0.999$ and $\epsilon = 10^{-8}$) in \cite{kingma2014adam} and used $10^{-3}$  for the value of alpha, the learning rate. In our experiment, Adam is drastically faster in training comparing to the classic SGD and AdaGrad. Figure \ref{fig:optimizers} shows average training convergence epochs for these four methods for three sampled users.

\section{The Experiment}
\label{sec:experiment}
In the experiment, we compare three methods that we discussed earlier with the proposed framework of this paper. The dataset is an anonymized CDR collection of 12 users in a period of 1.5-3 years. The summary of activities for each of the users is showed in Table \ref{tab:userdata}.

\begin{table}[ht]
	\caption{The summary of users activities in the dataset.}
	\label{tab:userdata}
	\centering
	\begin{tabular}{|c|c|c|c|c|c|}
		\hline \textbf{Name} & \textbf{Size of Records} & \textbf{Beginning Date} & \textbf{End Date} & \textbf{Period in Days} & \textbf{\# of Locations} \\ 
		\hline
		\hline USER01 & 39577 & 2014-03-20 & 2016-10-08 & 932 & 1337 \\
		\hline USER02 & 32225 & 2014-03-20 & 2017-01-09 & 1025 & 716 \\
		\hline USER03 & 29803 & 2014-03-20 & 2017-01-09 & 1025 & 2304 \\
		\hline USER04 & 26367 & 2014-03-20 & 2017-01-09 & 1025 & 942 \\
		\hline USER05 & 23999 & 2014-03-21 & 2017-01-09 & 1025 & 1532 \\
		\hline USER06 & 23626 & 2015-03-25 & 2017-01-02 & 648 & 1125 \\
		\hline USER07 & 15592 & 2014-03-23 & 2017-01-09 & 1023 & 1639 \\
		\hline USER08 & 13003 & 2015-03-20 & 2017-01-02 & 653 & 326 \\
		\hline USER09 & 8695 & 2014-04-21 & 2016-06-24 & 795 & 83 \\
		\hline USER10 & 8632 & 2014-03-27 & 2017-01-09 & 1018 & 305 \\
		\hline USER11 & 5061 & 2014-03-21 & 2017-01-08 & 1024 & 143 \\
		\hline USER12 & 3309 & 2015-07-14 & 2017-01-01 & 537 & 183 \\
		\hline
	\end{tabular}
\end{table}

We perform the experiment by splitting each user’s data to the train and test part with the corresponding proportion of 50\%-50\%. The first portion of the data is used as training data and the second half used as testing data. For the comparison, we use the average distance (stated in meters) between expected and predicted locations of the test data. We also perform grid search over the hyper-parameters of the proposed framework to show how the tuning of these parameters would affect the performance.

All the implementations are coded in R statistical and data analysis language \cite{Rlang}. We also use the RStudio software as the environment of coding, execution, debugging and visualization \cite{RStudio}. The implementation of the artificial neural networks done using the R interface to Keras \cite{chollet2015keras} with the backend engine of TensorFlow \cite{tensorflow2015}. We also use the ggplot2 \cite{ggplot2} package and ggmap \cite{ggmap} to plot the charts and maps of the results.  

\begin{figure}[h!]
	\centering
	\includegraphics[width=0.95\textwidth]{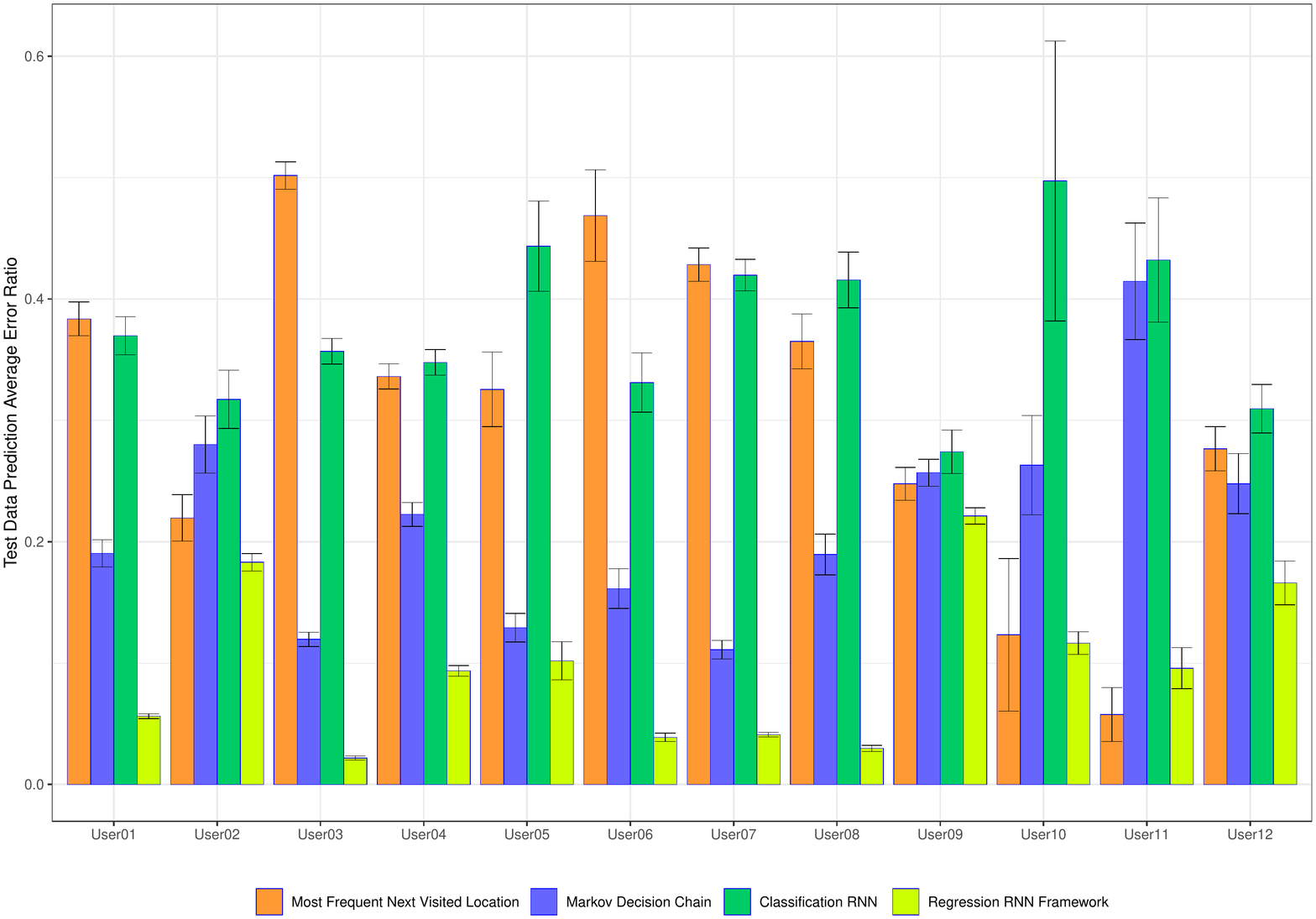}
	\caption{The ratio of the average error of three traditional models in comparison with the regression framework.}
	\label{fig:results}
\end{figure}

Figure \ref{fig:results} shows the overall average of errors in predicting next locations for the 12 users in our dataset using three existing methods and our proposed framework. The first method is the naïve model (Most Frequent Next Visited Location), the second one is Markov Decision Chain model, the third is the LSTM recurrent neural network classification model, and the fourth one is our proposed regression framework. It can be seen that the proposed framework outperforms the three traditional models in 11 cases, showing the average 55\% of the error decrease in respect to the average best-performed model.

We have stated the error values of each model for each of the 12 users in Table \ref{tab:resultsdetail}. The values of errors are in meter, stating the average distance between the predicted locations and the expected ones. These results are obtained using a grid search over the hyper-parameters of each model and the value of the error is stated for the best performance run.

\begin{table}[h!]
	\caption{The average error in meter for three traditional models and the regression framework.}
	\label{tab:resultsdetail}
	\centering
	\begin{tabular}{|c|c|c|c|c|}
		\hline \textbf{User} & \textbf{Most Frequent N.V.L.} & \textbf{Markov Chain} & \textbf{Classification RNN} & \textbf{Regression RNN FW} \\ 
		\hline
		\hline User01 & 3155 & 1566 & 3041 & 462 \\
		\hline User02 & 1637 & 2088 & 2366 & 1365 \\
		\hline User03 & 27058 & 6450 & 19248 & 1173 \\
		\hline User04 & 3546 & 2347 & 3668 & 987 \\
		\hline User05 & 6181 & 2452 & 8419 & 1934 \\
		\hline User06 & 10332 & 3556 & 7299 & 856 \\
		\hline User07 & 7178 & 1861 & 7033 & 688 \\
		\hline User08 & 12904 & 6698 & 14691 & 1051 \\
		\hline User09 & 646 & 670 & 715 & 577 \\
		\hline User10 & 4013 & 8559 & 16181 & 3791 \\
		\hline User11 & 843 & 6070 & 6326 & 1402 \\
		\hline User12 & 2115 & 1894 & 2366 & 1269 \\
		\hline
	\end{tabular} 
\end{table}

\begin{figure}[h!]
	\centering
	\includegraphics[width=0.7\textwidth]{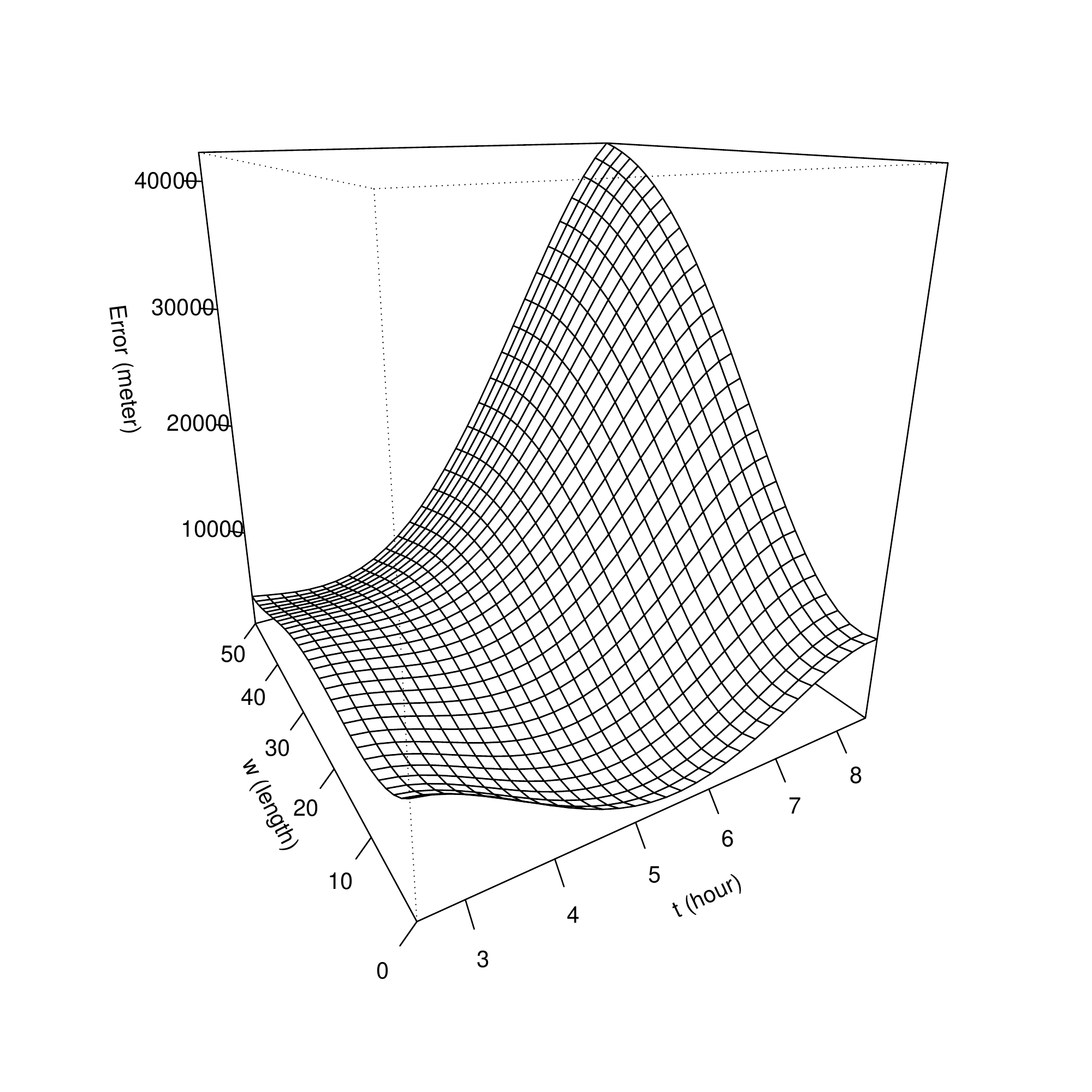}
	\caption{Grid search over the $t$ and $w$ parameters to find the global minimum of the prediction error for a user.}
	\label{fig:gridsearch}
\end{figure}

Two important parameters of the proposed framework are $t$, the factor of minimum user settlement timespan that has been used to slice a user’s trajectories into the smaller subsequences, and $w$ which is the length of the sliding window over the sequences and shows the maximum length of subsequences. As we discussed earlier, we expect that low values of $w$ result in low performance because of the scarce historical events we use in order to predict the next location. On the other hand, a large value for $w$ would cause under-performance because of the model overfitting the data. As we performed grid search over the values of $t$ and $w$, we obtain a 3-D plot where two horizontal axes show the values of $t$ and $w$, and the vertical axis contains the error values of the prediction. The result of the grid search for a sampled user is showed in Figure \ref{fig:gridsearch}. The result confirms the behavior that we discussed for the value of $w$. Optimal values of $w$ and $t$ should be obtained for each user independently.

\begin{figure}[h!]
	\centering
	\includegraphics[width=0.95\textwidth]{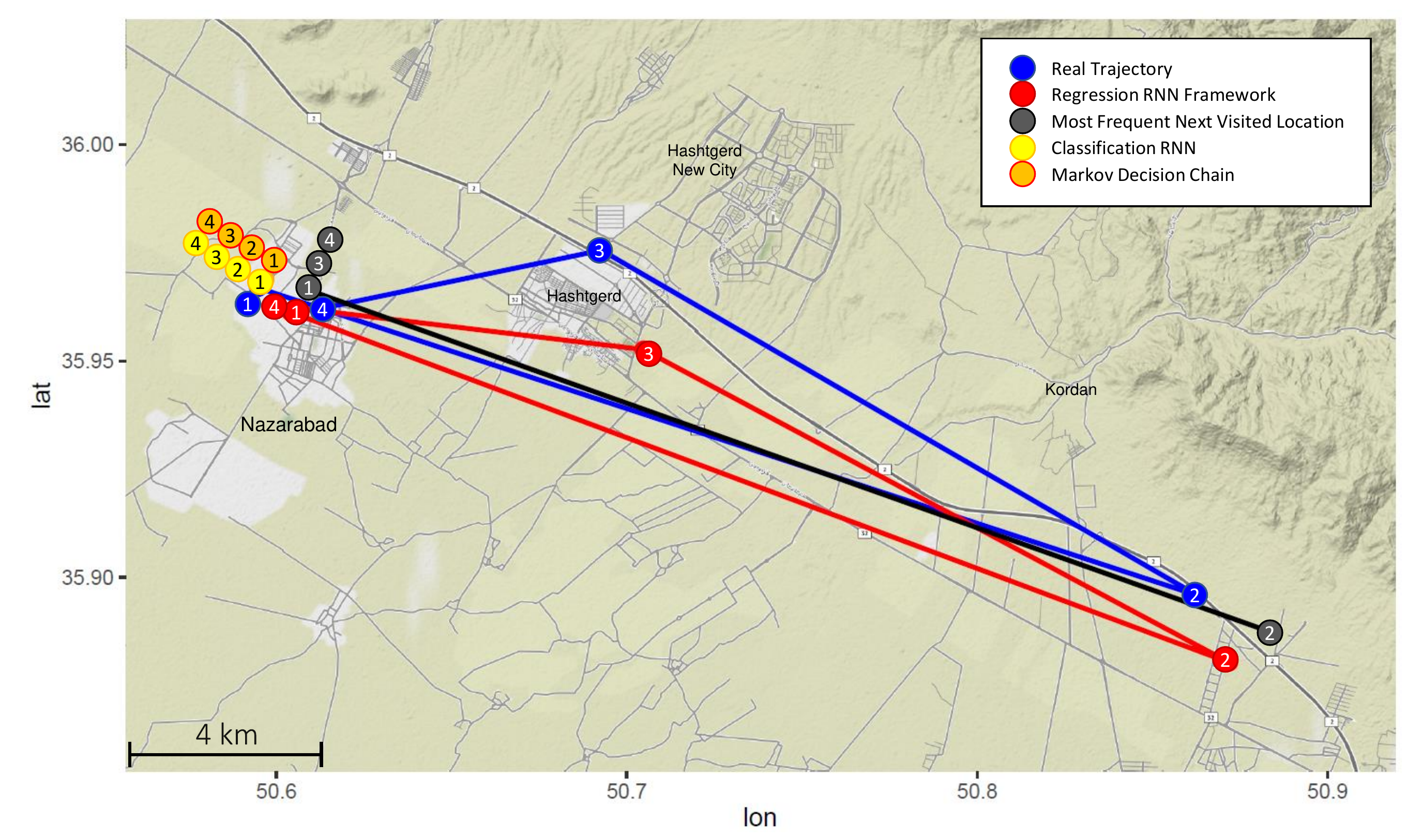}
	\caption{An example of predicting a trajectory of length 4 for a user with three traditional methods and the regression framework.}
	\label{fig:compare1}
\end{figure}

\begin{figure}[h!]
	\centering
	\includegraphics[width=0.95\textwidth]{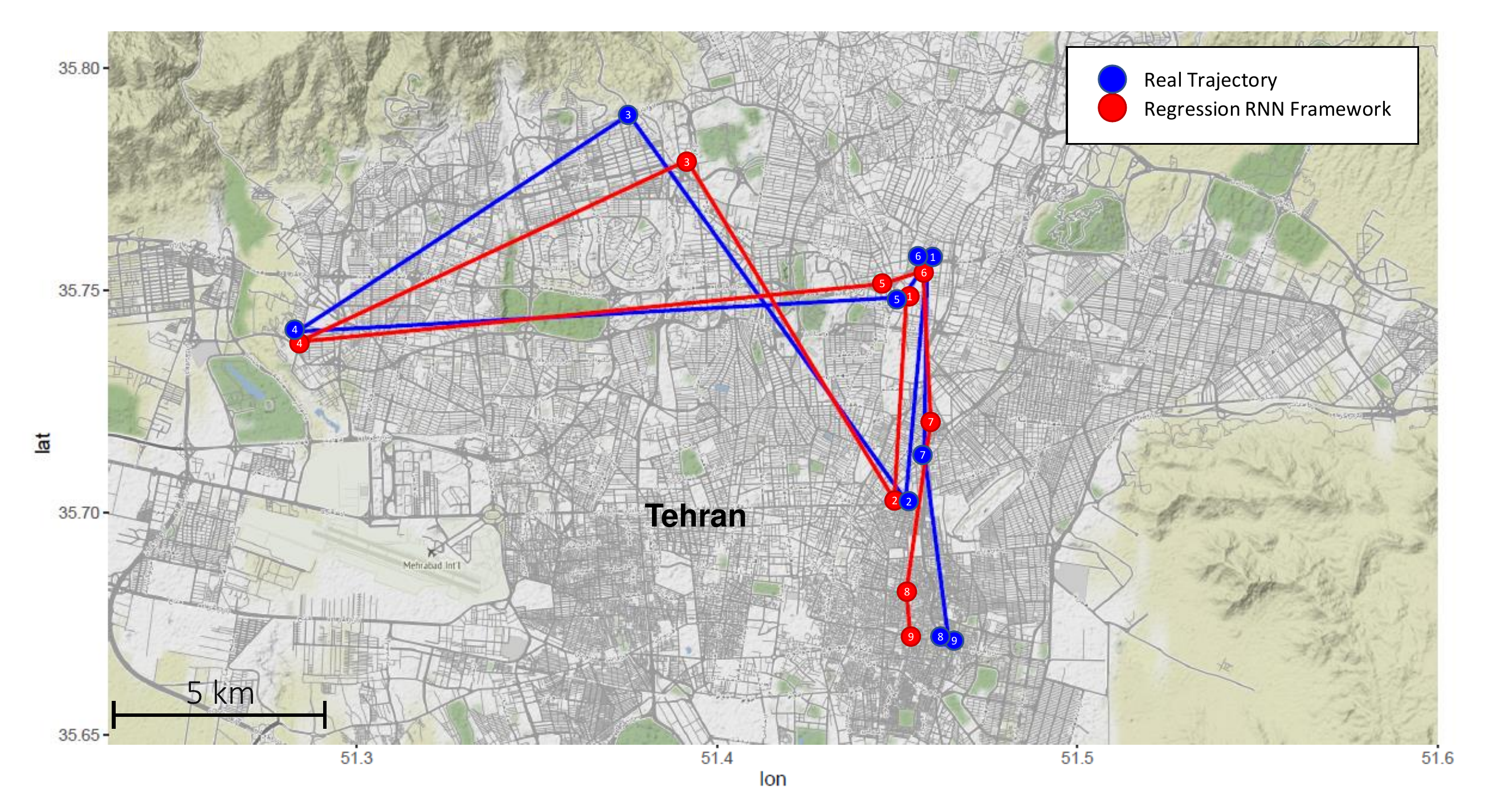}
	\caption{An example of predicting a trajectory of length 9 for a user with the regression framework.}
	\label{fig:compare2}
\end{figure}

Figure \ref{fig:compare1} illustrates the predictions of a trajectories with a length of 4 for a sampled user. As we discussed earlier, because of the classification process of traditional methods, their prediction result could not be a location that does not belong to the set of known locations of BTSes. In contrast, the prediction of our proposed regression framework could obtain location values anywhere along the geographical axes. Figure \ref{fig:compare1} shows that the regression model predicted the locations closely, but not the exact locations, but the result of the classification models are shifted to a distant location because of the limitation for the result to be in the location space of the user historical trajectories. Figure \ref{fig:compare2} also shows the predicted trajectory with the length of 9 for a user in compare to the real trajectory locations.

\begin{figure}[h!]
	\centering
	\includegraphics[width=0.95\textwidth]{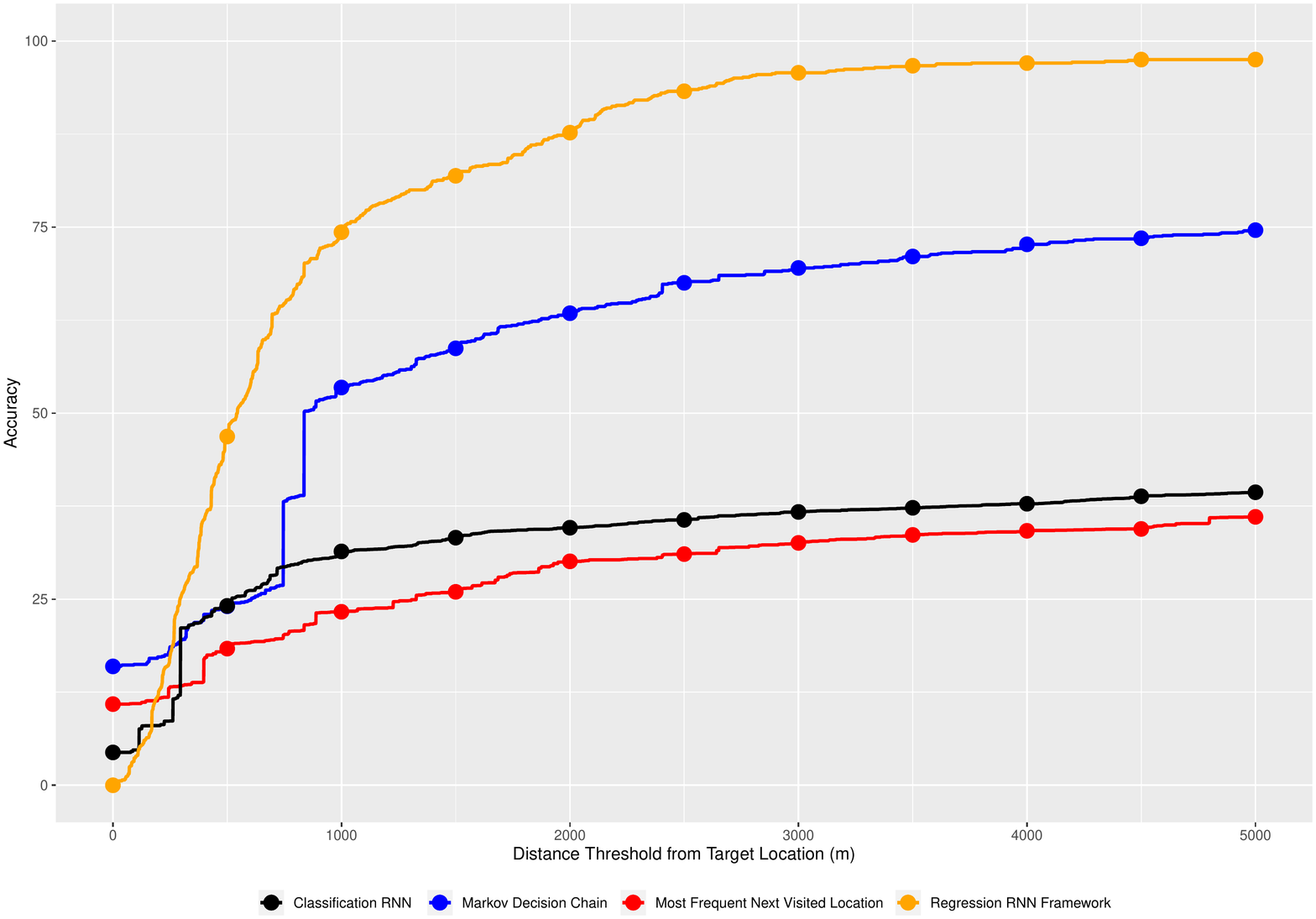}
	\caption{The value of accuracy vs. the minimum distance threshold between the target and the predicted location.}
	\label{fig:threshold}
\end{figure}

Although the performance of the regression model could not be measured by accuracy metrics, the error values of this model could be mapped to the accuracy metric. Figure \ref{fig:threshold} illustrates the relation between the accuracy and different values of the minimum distance thresold ($d$) of accepting prediction for the three traditional classification models and the proposed regression framework. The horizontal axis shows the distance threshold in meter and the vertical axis shows the accuracy value. We measure the accuracy of the model by assuming distances lower than $d$ from the expected location as a correct prediction. Now, using the definition of accuracy (i.e. number of correct predictions divided by the total number of predictions), we calculate the accuracy of the model for each value of $d$.

\section{Conclusion}
\label{sec:conclusion}
In this paper, we proposed a framework to predict the next location of users in a cellular network. This framework compensates the drawbacks of the sparsity in CDR data by a novel preparation method, resolves the limitations of traditional classification models for this problem, and proposes a unified process from raw CDR data to next location prediction. The proposed data preparation method is based on how cellular networks register their users’ trajectories. 
Another prominent contribution in this paper is proposing a regression model as the predictor in the framework. The regression model is a recurrent neural network based on two layers of LSTM nodes in order to catch hierarchical dependencies between adjacent locations and adjacent trajectories.  
Our experiments over 12 users CDR data over a period of 1.5-3 years shows that the proposed framework outperforms the traditional models by 74\% to 55\% of improvement in the error values. The reasons for the superiority of the proposed framework are our novel data preparation method and the specific design of the recurrent neural network model.


\end{document}